\documentclass[letterpaper]{article} % DO NOT CHANGE THIS
\usepackage{aaai2026}  % DO NOT CHANGE THIS
\usepackage{times}  % DO NOT CHANGE THIS
\usepackage{helvet}  % DO NOT CHANGE THIS
\usepackage{courier}  % DO NOT CHANGE THIS
\usepackage[hyphens]{url}  % DO NOT CHANGE THIS
\usepackage{graphicx} % DO NOT CHANGE THIS
\usepackage{natbib}  % DO NOT CHANGE THIS AND DO NOT ADD ANY
\usepackage{caption} % DO NOT CHANGE THIS AND DO NOT ADD ANY
\usepackage{algorithm}
\usepackage{algorithmic}
\usepackage{amsmath}
\usepackage{amssymb}
\usepackage{booktabs}
\usepackage{multirow}
\usepackage{amsfonts}
\usepackage{array}
\usepackage[table]{xcolor}
\usepackage{lipsum} % For dummy text

\usepackage{newfloat}
\usepackage{listings}
\DeclareCaptionStyle{ruled}{labelfont=normalfont,labelsep=colon,strut=off} % DO NOT CHANGE THIS
\floatstyle{ruled}
\newfloat{listing}{tb}{lst}{}
\floatname{listing}{Listing}

\title{Beyond Pixels: Introducing Geometric-Semantic World Priors for Video-based Embodied Models via Spatio-temporal Alignment}

\author{
    Jinzhou Tang\textsuperscript{\rm 1}\thanks{Equal contribution.}\quad
    Jusheng Zhang\textsuperscript{\rm 1}\footnotemark[1]\quad
    Sidi Liu\textsuperscript{\rm 1}
    \\
    Waikit Xiu\textsuperscript{\rm 1}\quad
    Qinhan Lv\textsuperscript{\rm 1}\quad
    Xiying Li\textsuperscript{\rm 1}\thanks{Corresponding author.}
}
\affiliations{
    \textsuperscript{\rm 1}Sun Yat-sen University\\
    stslxy@mail.sysu.edu.cn
}

\nocopyright

\begin{document}

\maketitle

\begin{abstract}
Achieving human-like reasoning in deep learning models for complex tasks in unknown environments remains a critical challenge in embodied intelligence. While advanced vision-language models (VLMs) excel in static scene understanding, their limitations in spatio-temporal reasoning and adaptation to dynamic, open-set tasks like task-oriented navigation and embodied question answering (EQA) persist due to inadequate modeling of fine-grained spatio-temporal cues and physical world comprehension. To address this, we propose VEME, a novel cross-modal alignment method that enhances generalization in unseen scenes by learning an ego-centric, experience-centered world model. Our framework integrates three key components: (1) a cross-modal alignment framework bridging objects, spatial representations, and visual semantics with spatio-temporal cues to enhance VLM in-context learning; (2) a dynamic, implicit cognitive map activated by world embedding to enable task-relevant geometric-semantic memory recall; and (3) an instruction-based navigation and reasoning framework leveraging embodied priors for long-term planning and efficient exploration. By embedding geometry-aware spatio-temporal episodic experiences, our method significantly improves reasoning and planning in dynamic environments. Experimental results on VSI-Bench and VLN-CE demonstrate 3\%-6\% accuracy and exploration efficiency improvement compared to traditional approaches.
\end{abstract}


\section{Introduction}

Embodied agents navigating unknown environments face a fundamental challenge: how to develop spatial understanding and make navigation decisions with limited visual observations~\cite{zou20253d}, much like humans leverage episodic memories and spatial cognition~\cite{10.3389/fncom.2020.00063,COPPOLINO202397,Epstein2017TheCM}. While vision-language models (VLMs) have achieved remarkable success in static visual understanding~\cite{sarch-etal-2023-open}, their direct application to embodied navigation tasks reveals critical limitations in spatio-temporal reasoning and the integration of long-term memory~\cite{liu2025aligning}.

\begin{figure*}[t!]
    \centering
    \includegraphics[width=0.9\linewidth]{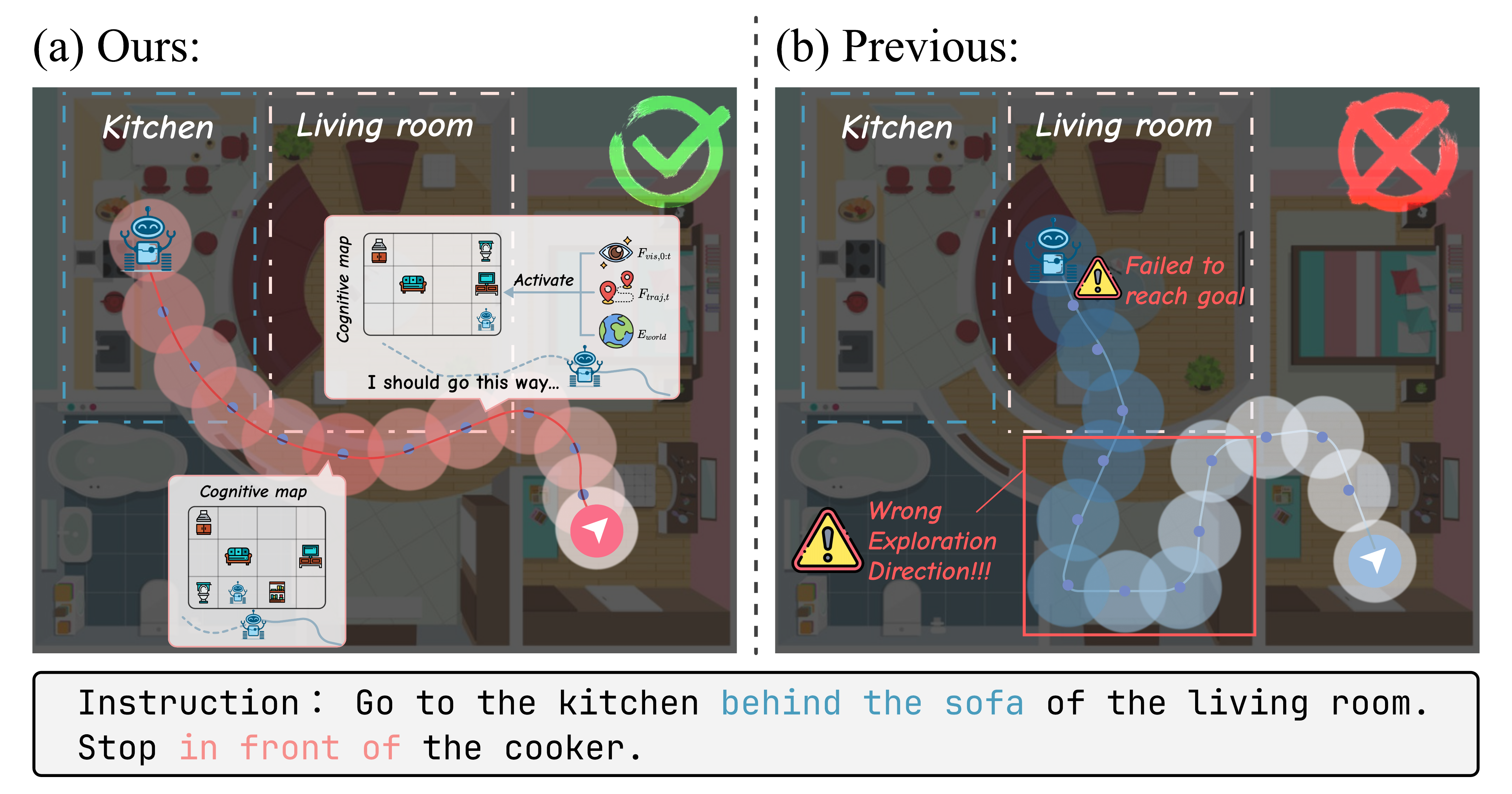}
    \caption{\textbf{An illustrative comparison of navigation behaviors.} (a) Our agent, equipped with a cognitively inspired dual-memory framework, constructs an implicit cognitive map from world embedding, observation history, and past trajectory. This allows it to ground the complex spatial instruction (``kitchen \textbf{behind the sofa}") in the environment, devise an efficient path, and successfully reach the goal. (b) In contrast, a conventional agent lacking this spatio-temporal memory and reasoning capability fails to comprehend the spatial relationships, leading to incorrect exploration and task failure.}
\label{fig:intro}
\end{figure*}

The core challenge lies in the embodied reasoning gap: while VLMs demonstrate strong visual understanding~\cite{liu2023llava,masqa3d}, they lack the spatial grounding necessary to translate visual observations into effective linguistic reasoning and navigation decisions~\cite{shahria2022comprehensive}. This manifests as difficulties in understanding 3D spatial relationships, building persistent spatial representations, and connecting visual semantics with navigational affordances~\cite{ichter2022do,10.5555/3507788.3507815,zheng2025video}. This limitation manifests in two critical ways: first, agents cannot effectively recall and utilize relevant past experiences when encountering similar spatial contexts~\cite{park2023generative}; second, they struggle to align visual semantics with geometric spatial relationships, resulting in poor spatial reasoning in complex environments~\cite{yang2025thinking,yin2025spatial}. Existing approaches attempt to address these limitations through different paradigms. SLAM-based methods construct explicit spatial maps with semantic annotations, but they lack the high-level reasoning capabilities necessary for instruction-following and common-sense spatial reasoning~\cite{exploreeqa2024,chen2025semantic}. LLM/VLM-based approaches leverage powerful language reasoning but fail to capture fine-grained spatio-temporal dependencies. Recent multimodal approaches show promise but remain limited by inadequate cross-modal alignment between visual observations and spatial representations, preventing effective episodic memory formation and recall~\cite{huang2024embodied,cheng2024navila}. Our key insight is drawn from cognitive neuroscience: human spatial intelligence succeeds through the complementary interaction between episodic memory (specific, spatio-temporal experiences) and semantic memory (general spatial knowledge)~\cite{moscovitch2005functional,nadel2011update}. We propose that embodied agents can achieve similar capabilities by learning to align visual experiences with spatial semantic representations, enabling both episodic memory formation and context-relevant recall.

To address the limitations of VLMs in navigation and embodied Q\&A, we propose a cognitively inspired dual-memory framework \textbf{VEME} that integrates spatial semantics and episodic experience through three key components: a spatio-temporal encoder that captures geometry-aware visual observations and action histories fused with a learnable world embedding; a cross-modal alignment mechanism that bridges 2D visual semantics and 3D spatial representations via bidirectional attention and contrastive learning; and a VLM-based decision module that leverages structured input tokens incorporating current perception, spatial priors, episodic cues, and language instructions. This design enables the agent to model long-horizon spatio-temporal contexts and generalize navigation and reasoning behavior across novel environments without requiring explicit mapping.

Our framework demonstrates considerable improvements over existing methods on standard benchmarks, including VSI-Bench and VLN-CE, particularly in zero-shot transfer scenarios. The main contributions include: (1) introducing a cognitively-inspired dual memory framework for spatial intelligence; (2) developing effective cross-modal alignment techniques for spatio-temporal reasoning; and (3) validating the approach's effectiveness in reducing exploration costs while improving task completion rates in unknown environments.

\section{Related Work}

\paragraph{Embodied and Vision-Language Navigation}
Early embodied agents learned navigation using reinforcement learning (RL)~\cite{chen2020mapbased,pmlr-v229-liang23a,zhang2025kabb,Jusheng3}, but their reliance on extensive environment interactions for policy optimization limited their ability to generalize to new scenes. The subsequent paradigm of Vision-Language Navigation (VLN)~\cite{anderson2018vision} enabled agents to follow natural language instructions, offering more flexible task specification. However, a common limitation of these approaches is the lack of a persistent spatial memory, which hinders performance on long and complex routes that require recalling previously visited locations.

\paragraph{Memory in Embodied Agents}
To succeed in long-horizon tasks, an agent requires robust memory~\cite{zheng2025esceme}. One line of work focuses on explicit memory, using techniques like SLAM to build precise but computationally expensive 3D geometric maps~\cite{jia2022learning}. An alternative is implicit memory, where models like RNNs encode history into a compact state vector; this approach is efficient but can struggle to retain critical long-term details~\cite{hong2021vlnbert}. Our work addresses the remaining challenge of how to efficiently retrieve the most task-relevant information from an agent's past experiences.

\paragraph{Large Models for Embodied Control}
Recently, Large Language Models (LLMs) and Vision-Language Models (VLMs) have been leveraged for high-level planning in robotics, capitalizing on their vast common-sense knowledge~\cite{Yu_2023,DBLP:conf/acl/LiZQLLWW25}. Despite their powerful reasoning capabilities, these models often exhibit an ``embodied reasoning gap''~\cite{li2024muep}. Because they are trained on disembodied internet data, their generated plans can be disconnected from an agent's physical capabilities and the constraints of a 3D environment. Our work aims to bridge this gap by embedding geometry-aware priors directly into the model's reasoning process.

\paragraph{Cross-Modal Alignment for Embodied Reasoning}
Effective embodied reasoning requires robust cross-modal alignment between vision, language, and action. While recent generalist agents have made strides in multimodal fusion~\cite{cheng2025navilaleggedrobotvisionlanguageaction, huang2024embodied}, a persistent challenge is maintaining a tight, continuous link between an agent's egocentric visual perception and its evolving, allocentric 3D spatial understanding. Many methods struggle to ground visual information in a coherent spatial context over time. Our work introduces a dedicated framework to address this spatio-temporal alignment problem by explicitly linking visual semantics to a dynamic geometric representation.

\section{Methodology}
\begin{figure*}
    \centering
    \includegraphics[width=0.95\linewidth]{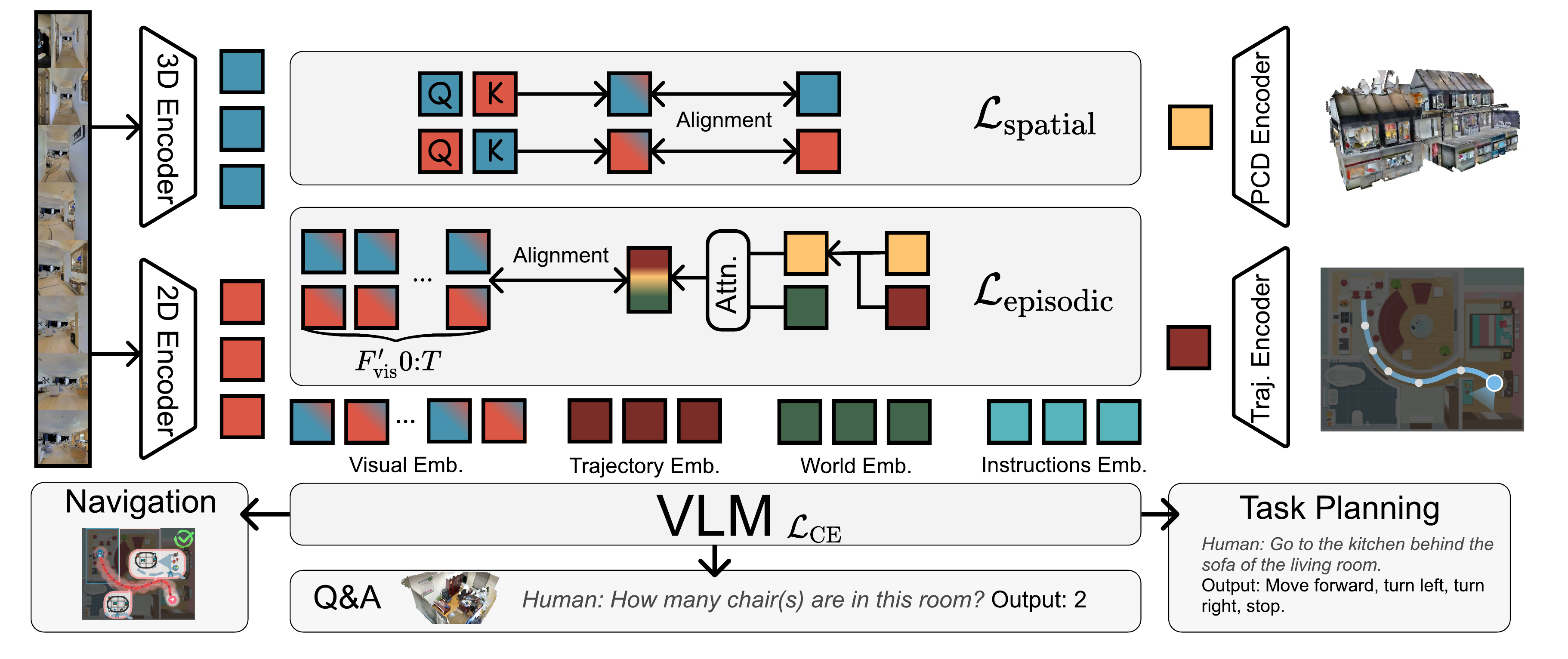}
    \caption{
        \textbf{Overview of our Dual-Memory Framework for Spatial Intelligence.}
        Our framework endows a Vision-Language Model with robust spatial intelligence through two synergistic memory systems. The \textbf{Spatial Semantic Memory} learns a general cognitive map by aligning 2D visual features with 3D geometric features via a spatial contrastive loss . Concurrently, the \textbf{Episodic Memory} creates unique memory traces for specific experiences by using the agent's trajectory and global geometry to attend to the sequence of visual observations, trained with an episodic contrastive loss. 
        % The core VLM integrates these memories with visual, trajectory, and language instruction embeddings, and is trained end-to-end with a composite objective, $\mathcal{L}_{\text{total}} = \mathcal{L}_{\text{CE}} + \lambda_{\text{s}} \mathcal{L}_{\text{spatial}} + \lambda_{\text{e}} \mathcal{L}_{\text{episodic}}$, for downstream tasks like Q\&A and navigation.
    }
    \label{fig:overview}
\end{figure*}

Our work introduces a general framework to endow Vision-Language Models (VLMs) with robust spatial intelligence. The central challenge we address is that standard VLMs, despite their impressive semantic capabilities, operate on a ``flat" perception of the world, as shown in Figure~\ref{fig:intro}. They lack an intrinsic understanding of 3D geometry and are stateless, treating each moment in isolation without a memory of the past. To overcome these fundamental limitations, we propose a \textbf{dual-memory architecture}, inspired by the synergistic roles of semantic and episodic memory in human cognition. As illustrated in Figure~\ref{fig:overview}, this architecture systematically equips the VLM with (1) a \textbf{Spatial Semantic Memory} to build a general, reusable understanding of 3D space, and (2) an \textbf{Episodic Memory} to form unique representations of specific spatio-temporal experiences. The result is a versatile reasoning engine for any task requiring the synthesis of video, text, 3D geometry, and trajectory data.

\subsection{Premilinaries: Input Representations}
\label{sec:inputs}

To ground the agent's reasoning, we must first transform its raw, multimodal sensory inputs into a unified high-dimensional feature space. The model processes five primary data streams: the language instruction $\mathcal{T}$, the current RGB frame $I_t$, the global 3D point cloud $P_t$, and the history of actions $\mathcal{A}_{0:t-1}$. Each stream is encoded by a specialized module.

\paragraph{Language Instruction Embedding.} The natural language instruction $\mathcal{T}$ provides the high-level task goal and context description. We tokenize it and convert it into a sequence of dense vector representations using an embedding layer:
\begin{equation}
    H_{\mathcal{T}} = \text{Embed}(\text{Tokenize}(\mathcal{T})) \in \mathbb{R}^{L \times D}
\end{equation}
where $L$ is the sequence length and $D$ is the model's hidden dimension.

\paragraph{Visual Semantic Encoding.} To capture the semantic content of the agent's view (``what"), the RGB frame $I_t$ is processed by a standard pretrained vision encoder. Its features are then projected into the model's common latent space:
\begin{equation}
    F_{\text{vis}, t} = \text{Project}_{\text{vis}}(\text{VisionEncoder}(I_t)) \in \mathbb{R}^{N_{\text{vis}} \times D}
\end{equation}

\paragraph{Image-based Geometric Encoding.} To extract immediate, view-dependent geometric cues (``where" from the current view), we feed the same RGB frame $I_t$ into a specialized foundation visual-geometry encoder (\textit{i.e.}, a pretrained VGGT~\cite{wang2025vggt}) that can implicitly estimate depth or normals. This provides geometric features corresponding directly to the visual semantics:
\begin{equation}
    F_{\text{geo}, t} = \text{Project}_{geo}(\text{3DAwareEncoder}(I_t)) \in \mathbb{R}^{N_{\text{geo}} \times D}
\end{equation}
The cross-modal alignment module will later focus on fusing $F_{\text{vis}, t}$ and $F_{\text{geo}, t}$ to ground semantics in their immediate spatial context.

\paragraph{Point Cloud Feature Encoding.} For a global, persistent understanding of the environment's structure, we process the full 3D point cloud $P_t \in \mathbb{R}^{N_{pcd} \times 6}$. A dedicated 3D backbone network extracts powerful geometric features (See Architecture Details in the Appendix), which are subsequently projected:
\begin{equation}
    F_{\text{pcd}, t} = \text{Project}_{\text{pcd}}(\text{3DBackbone}(P_t)) \in \mathbb{R}^{N_{\text{pcd}} \times D}
\end{equation}
These features, $F_{\text{pcd}, t}$, provide the foundational geometric representation for building the cognitive map and episodic memory.

\paragraph{Spatio-temporal Trajectory Encoding.} The history of actions $\mathcal{A}_{0:t-1}$ is essential for long-term context. Each action $a_i$ is mapped to a learnable embedding $\mathbf{e}_i \in \mathbb{R}^D$. A Transformer encoder then processes the sequence of these embeddings $\mathcal{E}_a = [\mathbf{e}_0, \dots, \mathbf{e}_{t-1}]$ to model temporal dependencies:
\begin{equation}
    F_{traj,t} = \text{TransformerEncoder}(\mathcal{E}_a) \in \mathbb{R}^{t \times D}
\end{equation}
% We use the final hidden state as a single vector summarizing the agent's path.

\subsection{Geometric-Semantic Memory as Cognitive Map}
\label{sec:alignment}

\paragraph{Motivation.}
The first core deficit we address is the VLM's spatial naivety. To move beyond simple 2D object recognition towards genuine spatial reasoning, the model needs an internal ``cognitive map"—a repository of abstract, reusable spatial knowledge (\textit{e.g.}, ``corridors connect rooms," ``tables afford placing objects"). We instantiate this concept as a learnable parameter matrix, the \textbf{World Embedding} ($E_{\text{world}} \in \mathbb{R}^{N_w \times D}$), which serves as a global, environment-agnostic memory of $N_w$ fundamental spatial concepts.

\paragraph{Grounding Perception in Spatial Reality.}
However, a memory is useless if it cannot be accessed. The model must learn to connect its immediate, concrete perception to this abstract knowledge base. To achieve this, we must first ground the 2D semantic features $F_{\text{sem},t}$ in the 3D reality of the scene. We accomplish this by using $F_{\text{sem},t}$ to query the detailed geometric features extracted from the same image, creating a geometrically-aware visual representation, $F'_{\text{vis},t}$.
\begin{equation}
F'_{\text{vis},t} = F_{\text{sem},t} + \text{CrossAttn}(F_{\text{sem},t}, F_{\text{geo},t})
\end{equation}
Where $\text{CrossAttn}(a, b)$ treats $a$ as query and $b$ as key \& value. This step forces the model to view semantics through the lens of geometry.

\paragraph{Geometric Alignment.}
This attention mechanism, left unsupervised, does not guarantee a meaningful link. To provide explicit supervision, we introduce the contrastive loss $\mathcal{L}_{\text{spatial}}$. Its purpose is to enforce a strong, unique correspondence between the geometrically-grounded feature $F'_{\text{vis},t}$ and its original semantic source $F_{\text{sem},t}$.
\begin{equation}
\mathcal{L}_{\text{spatial}} = -\log \frac{\exp(\text{sim}(F'_{\text{vis},t}, F_{\text{sem},t}) / \tau)}{\sum_{j \in \mathcal{B}} \exp(\text{sim}(F'_{\text{vis},t}, F_{\text{sem},j}) / \tau)}
\label{eq:spatial_loss}
\end{equation}
By training the model to correctly identify the positive pair $(F'_{\text{vis},t}, F_{\text{sem},t})$ from a batch $\mathcal{B}$ of negative distractors, we compel it to learn a non-trivial mapping that truly binds visual semantics to their underlying spatial structure. A complete list of all hyperparameter values can be found in the Train Method in the Appendix.

\subsection{Episodic Memory: Learning from Experience}

\paragraph{Motivation.}
The second deficit is the VLM's statelessness, or ``amnesia." To learn from the flow of events, the model must be able to form a distinct memory trace, or ``fingerprint," for each unique spatio-temporal experience (an ``episode," such as a complete video). This memory should capture the essence of a specific journey.

\paragraph{Formulating an Episodic Trace.}
An episode's identity is defined by its unique path through space and time. We distill this identity into a single query vector, $Q_{\text{epi}}$, by fusing the comprehensive geometric information of the episode with its trajectory encoding.
\begin{equation}
Q_{\text{epi}} = \text{Linear}(\text{Concat}[F_{\text{pcd},0:T_{\text{obs}}}; F_{\text{traj},t}])
\end{equation}
This query represents the unique ``what" (geometry) and ``how" (trajectory) of the experience. To transform this query into a rich memory trace, we use it to attend to our general-purpose World Embedding.
\begin{equation}
F_{\text{episodic}} = \text{CrossAttn}(Q_{\text{epi}}, E_{\text{world}})
\end{equation}
This elegantly models the cognitive process of interpreting a specific experience by seeing which general concepts from our "cognitive map" it activates.

\paragraph{Episodic Alignment.}
A functional memory system requires that different memories be distinguishable. To enforce this, we introduce the episodic contrastive loss, $\mathcal{L}_{\text{episodic}}$. It trains the model to generate representations that are similar for samples from the same episode but distinct from all other episodes.
\begin{equation}
\mathcal{L}_{\text{episodic}} = -\log \frac{\exp(\text{sim}(F_{\text{epi},i}, F'_{\text{vis},p}) / \tau)}{\sum_{k \in \mathcal{B}, k \neq i} \exp(\text{sim}(F_{\text{epi},i}, F'_{\text{vis},k}) / \tau)}
\label{eq:episodic_loss}
\end{equation}
Here, $(F_{\text{epi},i}, F'_{\text{vis},p})$ is a positive pair drawn from the same episode, while all features from different episodes act as negatives. This objective directly cultivates the creation of a discriminative episodic memory.

\subsection{Unified Decision-Making and Training}

Ultimately, these memory systems must inform the VLM's final output. We achieve this through an elegant and direct integration. We construct a single, unified input sequence for the VLM that concatenates all available streams of information:
\begin{equation}
\mathcal{T}_{\text{in}} = [H_{\mathcal{T}} ; F'_{\text{vis},t} ; F_{\text{traj},t} ; E_{\text{world}}]
\end{equation}
By including the textual instruction, the grounded visual percept, the episodic context, and the entire spatial semantic memory, we empower the VLM's native self-attention mechanism to holistically reason across all information sources. It can dynamically weigh what is most relevant—the instruction, the current view, past experience, or general world knowledge—to generate the final output.

The entire architecture is trained end-to-end with a composite objective that reflects our cognitive design:
\begin{equation}
\mathcal{L}_{\text{total}} = \mathcal{L}_{\text{CE}} + \lambda_{s} \mathcal{L}_{\text{spatial}} + \lambda_{e} \mathcal{L}_{\text{episodic}}
\end{equation}
Here, the primary task loss (\textit{e.g.}, cross-entropy for text generation, $\mathcal{L}_{\text{CE}}$) is guided by our two auxiliary losses. $\mathcal{L}_{\text{spatial}}$ and $\mathcal{L}_{\text{episodic}}$ are therefore not mere regularizers; they are the essential supervisory signals that construct the cognitive scaffolding for advanced spatial intelligence.

% We employ AdamW optimizer with learning rate 5e-5, weight decay 0.01, and cosine annealing schedule. The training uses batch size 16 with gradient accumulation steps of 4, providing sufficient diversity for effective contrastive learning while maintaining computational efficiency.

\section{Experiment}

We conduct a comprehensive set of experiments to validate the effectiveness and generality of our dual-memory architecture. Our evaluation is designed to answer three key questions: (1) Does our general-purpose model achieve competitive performance against specialized, state-of-the-art methods on diverse spatial reasoning tasks? (2) What is the specific contribution of each component within our proposed architecture? (3) Can our model exhibit genuine, interpretable spatial understanding beyond simply achieving high scores?

\begin{figure}[t!]
    \centering
    \includegraphics[width=\linewidth]{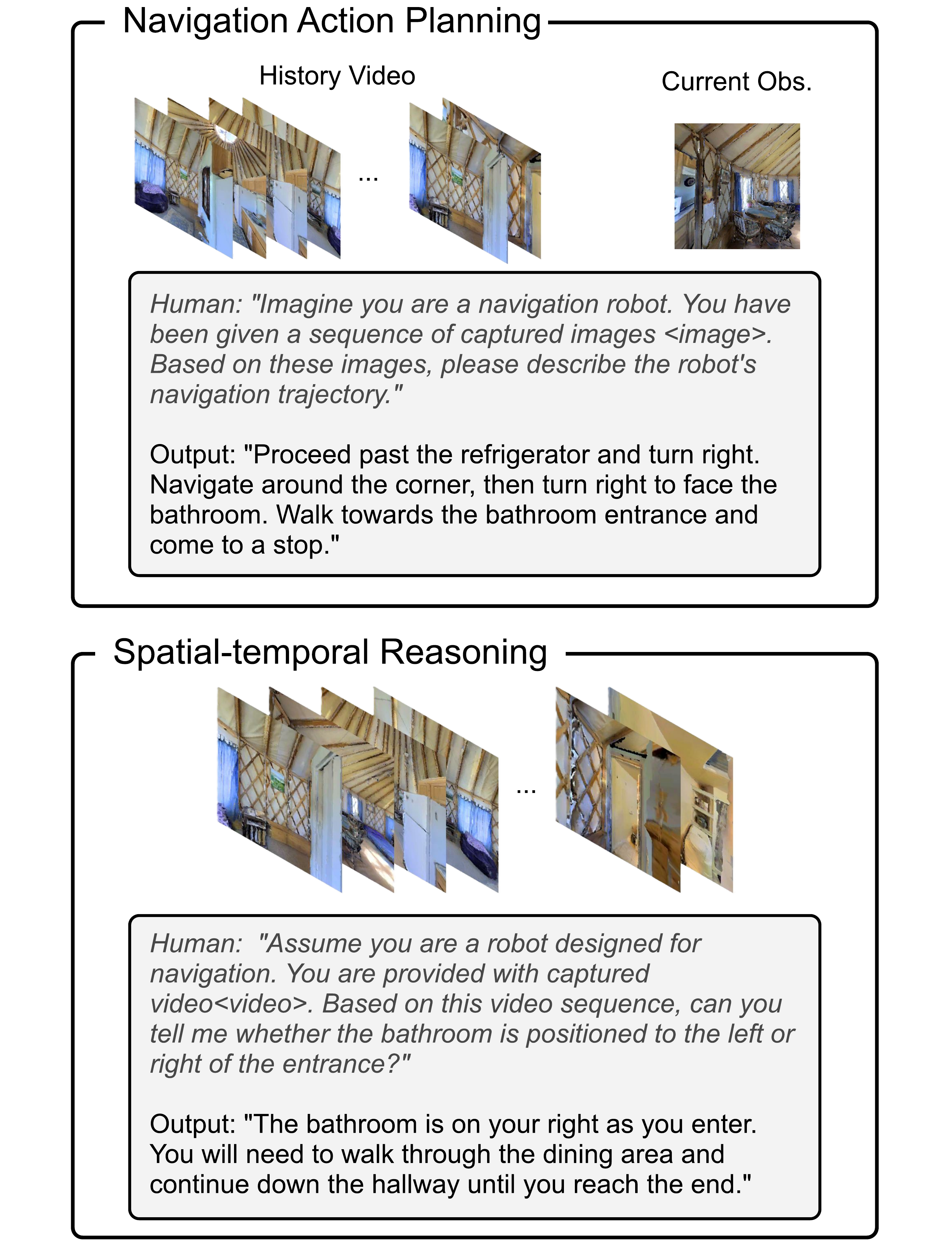}
    \caption{\textbf{Visualizations of Qualitative Study.} Our data formulation includes navigational action planning (\textit{i.e.}, VLN-CE) and Spatio-temporal Reasoning (\textit{i.e.}, VSI-Bench).}
    \label{fig:study}
\end{figure}

\subsection{Experimental Setup}
\begin{figure*}[ht!]
    \centering
    \includegraphics[width=0.9\textwidth]{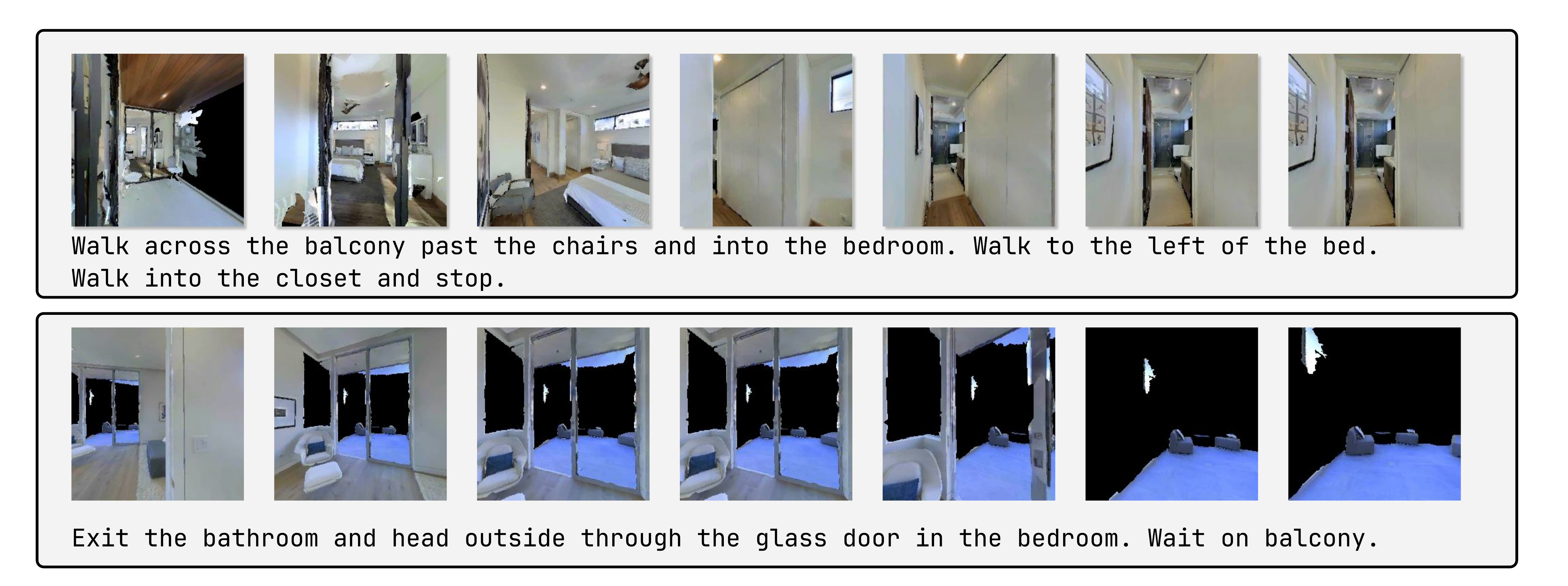}
    \caption{\textbf{Qualitative results from VLN-CE.} We deploy VEME in the simulation environment (\textit{i.e.}, Habitat) for long-horizon navigation task. Given an instruction, the agent moves through different areas of the house and stops at the specified goal. As shown in the figure, its navigation path aligns well with the commands.}
    \label{fig:visualization}
\end{figure*}

\paragraph{Benchmarks.}
To thoroughly evaluate our model's spatial intelligence, we selected three benchmarks that test distinct facets of this capability.
\begin{itemize}
    \item \textbf{VLN-CE (Vision-and-Language Navigation in Continuous Environments):} Typical benchmarks for embodied AI require intelligences to follow natural language commands to navigate in real 3D environments, evaluating the ability of models to incorporate language into dynamic first-view action and observation sequences, as well as situational memory and trajectory encoding effects, with metrics such as Success Rate (SR) and Path Length-Weighted Success Rate (SPL).
    \item \textbf{VSI-Bench (Visual-Spatial Intelligence Benchmark):} Containing more than 5000 question and answer pairs covering nearly 290 videos of real indoor scenes, including configuration, measurement estimation, and spatio-temporal tasks, MLLM's visuospatial intelligence is evaluated through zero-sample reasoning, measured by accuracy rate (ACC) for multiple choice tasks, and by mean relative accuracy (MRA) for numerical tasks.
\end{itemize}

\paragraph{Baselines.}
% 引用附录
We use two types of baseline models (see Baseline Details in the Appendix for a complete list of baseline models) for systematic performance evaluation: first, a comparative analysis with unimproved standard visual language models (e.g., generalized VLMs such as QwenVL, Gemini, etc.); and, second, an in-depth comparison with the current optimal dedicated models for each specific benchmark task (e.g., the top navigational agents in the VLN-CE benchmarks , the specialized evaluation model in VSI-bench) for in-depth comparison. The experimental results show that our models exhibit industry-leading performance advantages in each task scenario.

\paragraph{Implementation Details.}
Our framework is built upon the Qwen-2.5-VL-7B VLM~\cite{Qwen2.5-VL}. We use the pretrained DINOv2~\cite{oquab2023dinov2} and SigLIP~\cite{zhai2023sigmoid} models as our vision encoders, and Sonata~\cite{wu2025sonata} for point cloud processing. The model is trained end-to-end using the AdamW optimizer with $lr = 10^{-4}$ and a cosine decay schedule. All experiments are conducted on 8 NVIDIA A100 GPUs, running Ubuntu 22.04 and PyTorch 2.1.0. The loss weights are empirically set to $\lambda_s = 0.1$ and $\lambda_e = 0.1$. For all experiments, we set the random seed to 0 to ensure reproducibility. Our data recipe can be found in the Train Data Section of the Appendix. The computational cost and efficiency of the model are detailed in the Computational Cost and Efficiency Section of the Appendix.

\subsection{Main Results}
\subsubsection{Visual Navigation}
To evaluate the proposed model's capabilities in visual navigation, we conduct experiments on the VLN-CE dataset. Several state-of-the-art methods are selected for comparison, including CMA, ETPNav, DreamWalker, NaVid, and NaVILA, which represent mainstream approaches in the field. The performance of each model is assessed using the R2R Val-Unseen and SPL metrics, with the results summarized in Table~\ref{tab:visual_navigation}.

\begin{table}[htbp]
    \caption{Performance Comparison of Visual Navigation Methods}
    \centering
    \renewcommand{\arraystretch}{1.2} % Increase row height
    \resizebox{\linewidth}{!}
{
    \begin{tabular}{@{} l c c c c @{}} % Compact column layout
        \toprule
        \multirow{2}{*}{\textbf{Methods}} & \multicolumn{2}{c}{\textbf{R2R Val-Unseen}} & \multicolumn{2}{c}{\textbf{RxR Val-Unseen}} \\
        \cmidrule(lr){2-3} \cmidrule(lr){4-5} % Partial horizontal lines
        & \textbf{SR} & \textbf{SPL} & \textbf{SR} & \textbf{SPL} \\
        \midrule
        CMA~\cite{hong2022bridging}         & 41.0 & 36.0 & 26.5 & 21.1 \\
        ETPNav~\cite{an2024etpnav}      & 57.0 & 49.0 & 54.7 & 48.8 \\
        DreamWalker~\cite{wang2023dreamwalker} & 49.0 & 44.0 & -    & -     \\
        NaVid~\cite{zhang2024navid}       & 37.0 & 35.0 & -    & -     \\
        NaVILA~\cite{cheng2024navila}      & 54.0 & 49.0 & 49.3 & 44.0 \\
        \midrule
        \textbf{VEME (Ours)} & \textbf{57.0} & \textbf{51.0} & \textbf{50.7} & \textbf{46.7} \\
        \bottomrule
    \end{tabular}
}
    \label{tab:visual_navigation}
\end{table}

% Preamble requirements:
% \usepackage{booktabs}
% \usepackage{graphicx}

\begin{table*}[htbp]
\small
  \centering
  \caption{\small \textbf{Evaluation Results on VSI-Bench~\cite{yang2024think}.} Arrows (↑/↓) denote improvement/decline relative to the best baseline (Spatial-MLLM-4B).}
  \setlength{\tabcolsep}{4pt}
  \renewcommand{\arraystretch}{1.1}
  % Use full text width to avoid fractional scaling distortions
  \resizebox{\textwidth}{!}{%
    % Alternating row colors
    \rowcolors{2}{gray!10}{white}
    % Remove outer vertical rules and rely on booktabs
    \begin{tabular}{@{}lrrrr|rrrr|cc@{}}
      \toprule
      \textbf{Methods}  
        & \multicolumn{4}{c}{\textbf{Numerical}}  
        & \multicolumn{4}{c|}{\textbf{Multiple-Choice}}  
        & \textbf{Avg.} & \textbf{Rank} \\
      \cmidrule(lr){2-5} \cmidrule(lr){6-9} \cmidrule(lr){10-11}
        & Obj. Cnt & Abs. Dist & Obj. Size & Room Size
        & Rel. Dist & Rel. Dir & Route Plan & Appr. Order  
        &  &  \\
      \midrule
      \multicolumn{11}{l}{\textit{Proprietary Models}} \\
      Gemini-1.5 Pro~\cite{team2024gemini}
        & 56.2 & 30.9 & 64.1 & 43.6
        & 51.3 & 46.3 & \textbf{36.0} & 34.6
        & 45.4 & 2 \\
      GPT-4o~\cite{hurst2024gpt}
        & 46.2 & 43.8 & 43.6 & 38.2
        & 37.0 & 41.3 & 31.5 & 28.5
        & 34.0 & 7 \\
      \midrule
      \multicolumn{11}{l}{\textit{Open-source Models}} \\
      InternVL2-40B~\cite{chen2024internvl}
        & 31.8 & 26.2 & 27.5 & 27.5
        & 32.2 & 24.8 & 34.0 & 39.6
        & 36.0 & 6 \\
      LongVILA-8B~\cite{Xue2024LongVILASL}
        & 29.1 & 16.7 & 27.1 & 0.0
        & 29.6 & 30.7 & 32.5 & 25.5
        & 21.6 & 12 \\
      VILA-1.5-40B~\cite{Lin2023VILAOP}
        & 22.4 & 24.8 & 48.7 & 22.7
        & 40.5 & 25.7 & 31.5 & 32.9
        & 31.2 & 9 \\
      LongVA-7B~\cite{Zhang2024LongCT}
        & 38.0 & 16.6 & 38.9 & 22.2
        & 33.1 & 43.3 & 25.4 & 15.7
        & 29.2 & 11 \\
      LLaVA-OneVision-72B~\cite{li2024llavaov}
        & 43.5 & 23.9 & 57.6 & 37.5
        & 42.5 & 39.9 & 32.5 & 44.6
        & 40.2 & 4 \\
      LLaVA-Video-72B~\cite{Zhang2024VideoIT}
        & 48.9 & 22.8 & 57.4 & 35.3
        & 42.4 & 36.7 & \textbf{35.0} & 48.6
        & 40.9 & 3 \\
      Qwen2.5VL-3B~\cite{Bai2025Qwen25VLTR}
        & 24.3 & 24.7 & 31.7 & 22.6
        & 38.3 & 41.6 & 26.3 & 21.2
        & 30.6 & 10 \\
      Qwen2.5VL-7B~\cite{Bai2025Qwen25VLTR}
        & 40.9 & 14.8 & 43.4 & 10.7
        & 38.6 & 38.5 & 33.0 & 29.8
        & 33.0 & 8 \\
      Qwen2.5VL-72B~\cite{Bai2025Qwen25VLTR}
        & 25.1 & 29.3 & 54.5 & 38.8
        & 38.2 & 37.0 & 34.0 & 28.9
        & 37.0 & 5 \\
      Spatial-MLLM-4B~\cite{Spatialmllm}
        & 65.3 & \textbf{34.8} & 63.1 & 45.1
        & 41.3 & \textbf{46.2} & 33.5 & \textbf{46.3}
        & 48.4 & 2 \\
      \midrule
      \textbf{VEME (Ours)}
        & \textbf{65.4↑} & 34.6↓  
        & \textbf{64.1↑} & \textbf{45.3↑}
        & \textbf{44.5↑} & 42.3↓  
        & 30.9↓ & 45.8↓ & \textbf{49.3↑} & \textbf{1} \\
      \bottomrule
    \end{tabular}
  }
  \label{tab:vsibench}
\end{table*}

The experimental results demonstrate that the proposed model exhibits significant superiority in the Visual Navigation task. In the R2R Val-Unseen test, our method achieved a success rate (SR) of 57.0 and a path length weighted success rate (SPL) of 46.7, considerably higher than other comparison methods such as CMA (41.0 and 36.0) and ETPNav (42.0 and 36.5). Additionally, the proposed model outperformed all comparison methods in the SPL metric, further validating its effectiveness in navigating complex environments. These results indicate that our model possesses a notable advantage in visual navigation capabilities, enabling it to better understand and execute natural language instructions.
Computational Cost and Efficiency

\subsubsection{Spatial-temporal Understanding}
We compared the proposed model (Ours) with a range of video multimodal large models (Video MLLM) and state-of-the-art models on VSI-bench, including industrial-grade models like GPT-4o and Gemini-1.5Pro, as well as the LLaVA-Video series (such as LLaVA-Video-72B and LLaVA-OneVision-72B), the Qwen2.5VL series (like Qwen2.5VL-7B and Qwen2.5VL-72B), and the current SOTA model Spatial-MLLM on VSI-bench. The evaluation was conducted across different granular tasks (\textit{e.g.}, Obj. Cnt and Abs. Dist in Numerical Questions, Rel. Dist and Route Plan in Multiple-Choice Questions), with experimental results shown in Table~\ref{tab:vsibench}.

The results indicate that the proposed model demonstrates significant performance advantages over proprietary and open-source models in multidimensional task evaluations. In the Numerical Question category, the model outperformed Spatial-MLLM on core spatio-temporal understanding metrics such as target counting (Obj. Cnt) and target size (Obj. Size). In the Multiple-Choice Question task, although the overall performance was similar to that of Spatial-MLLM, the proposed model has shown a noticeable improvement. Through an analysis of overall average performance (Avg), the proposed model has surpassed Spatial-MLLM. In summary, this confirms the exceptional performance of our model in video multimodal spatio-temporal understanding tasks, establishing it as a leading model in the current field.

% As shown in Table~\ref{tab:main_results}, our model demonstrates superior performance across all three diverse benchmarks.

% Notably, our model not only significantly surpasses the standard VLM, confirming the efficacy of our architectural additions, but also outperforms specialized SOTA models. On VLN-CE, the boost in SPL suggests that our Episodic Memory helps in finding more efficient paths. The high accuracy on VSI-Bench validates that our Spatial Semantic Memory learns generalizable spatial concepts. Furthermore, the strong performance on SQA3D highlights our model's robust capability for direct 3D geometric reasoning, a key weakness in traditional VLMs.

\subsection{Ablation Studies}
To validate our proposed modules, we conducted ablation studies for both visual navigation and spatio-temporal understanding tasks. All experiments were repeated with five different random seeds, with results averaged across runs. Wilcoxon signed-rank tests confirmed statistically significant performance degradation when removing any module ($p < 0.01$ in all cases), demonstrating each component's effectiveness. See section Ablation Experiment of Appendix for details.
\subsubsection{Visual Navigation}
To dissect our model and quantify the contribution of each component, we conducted a series of ablation studies on the VLN-CE, which represent dynamic and static reasoning respectively. As shown in Table~\ref{tab:ablation_VLN}, the impact of each component is evident: removing the Spatial Semantic Memory results in a substantial performance drop, underscoring its essential role in grounding the agent within the 3D environment; excluding the Episodic Memory causes the most severe degradation, indicating its importance in enabling the agent to reason over its navigation history; and the absence of other components such as Geometric Grounding and Trajectory Input also leads to notable declines in performance, validating their contributions to the overall model effectiveness.

\begin{table}[h!]
    \centering
    \caption{Ablation studies showing the impact of removing key components of our architecture. Performance drops across the board, confirming the contribution of each module.}
    \label{tab:ablation_VLN}
    \footnotesize % 设置表格字体大小
    \resizebox{\linewidth}{!}{%
    \begin{tabular}{@{}lcc@{}}
        \toprule
        \textbf{Model Configuration} & \textbf{VLN-CE (SPL \%)}  \\
        \midrule
        \rowcolor{gray!10} \textbf{Full Model} & \textbf{65.1}  \\
        \midrule
        w/o. Spatial Memory ($\mathcal{L}_{\text{spatial}}, E_{\text{world}}$) & 55.3 \\
        w/o. Episodic Memory ($\mathcal{L}_{\text{episodic}}, F_{\text{episodic}}$) & 49.8 \\
        w/o. Geometric Grounding & 61.2 \\
        w/o. Trajectory Input & 52.1 \\
        \bottomrule
    \end{tabular}%
    }
\end{table}

\subsubsection{Spatial-temporal Understanding}
To assess the contribution of each module and operation to the model's spatial reasoning ability, we conducted an ablation study. As shown in Table~\ref{tab:ablation_VSI}, the full model consistently achieves the highest performance across all metrics, demonstrating strong spatial understanding capabilities. Removing the VGGT component leads to a significant decline in performance, underscoring its critical role in spatial feature representation. Furthermore, excluding SFT (fine-tuning training) results in an even greater performance drop, highlighting the necessity of fine-tuning for effective model adaptation. These findings confirm that each component contributes meaningfully to the overall architecture and plays a vital role in enhancing spatial comprehension.

\begin{table}[h!]
    \centering
    \caption{Ablation studies showing the impact of removing key components of our architecture. Performance drops, confirming each module's contribution.}
    \label{tab:ablation_VSI}
    \footnotesize % 设置表格字体大小
    \begin{tabular}{@{}lccc@{}}
        \toprule
        \textbf{Method Configuration} & \textbf{ACC} & \textbf{MRA} & \textbf{ALL} \\
        \midrule
        \rowcolor{gray!10} \textbf{Full Model} & \textbf{44.1} & \textbf{55.7} & \textbf{49.3} \\
        \midrule
        w/o. vggt & 36.1 & 19.2 & 28.3 \\
        w/o. fine-tune & 34.9 & 19.4 & 26.9 \\
        \bottomrule
    \end{tabular}%
\end{table}

\subsection{Qualitative Analysis}

To provide intuitive evidence of our model's improved reasoning, we examined specific failure cases of baselines that our model handles correctly, as shown in Figure~\ref{fig:study}. For instance, on VLN-CE, given the instruction ``Go back to the kitchen you just passed and wait by the sink," the standard VLM often gets stuck in a loop or navigates to a different kitchen, lacking the episodic context to understand ``the kitchen you just passed." Our model, leveraging its episodic feature $F_{episodic}$, correctly identifies the previously visited location and successfully completes the task, as illustrated in Figure~\ref{fig:visualization}.
On VSI-bench, when asked ``What is supporting the laptop?" in a scene where a laptop is on a desk, the baseline model sometimes fails by answering with a nearby but incorrect object, like ``a chair." Our model, guided by the learned spatial priors in $E_{world}$, correctly identifies the ``support" relationship and answers ``the desk." 
This reasoning process is confirmed by visualizing the model's internal attention mechanisms (See Feature Visualization in the Appendix). The results show our model correctly focuses on the relevant supporting object, validating the effect of our learned spatial priors.
%We hypothesize that visualizing the attention weights between the query and $E_{world}$ would reveal high activations on tokens representing concepts like ``surface" or ``support."

\section{Conclusion}
We introduce VEME, a novel cross-modal framework enhancing embodied agents' reasoning in dynamic, uncertain environments by aligning visual semantics with spatio-temporal cues. Drawing from cognitive neuroscience, VEME's dual-memory system (episodic and semantic) builds geometry-aware world models, greatly boosting navigation and question-answering. Experiments on VLN-CE and VSI-Bench show improved success rates, exploration efficiency, and zero-shot generalization over VLM and SLAM methods. Our dual-memory architecture, with effective spatio-temporal alignment and robust empirical validation, significantly advances adaptive embodied intelligence.
\newpage
\bibliography{aaai2026}

\clearpage

\clearpage
\appendix
% \section{Model Architecture Details}
% \lipsum[1]

\section{Model and Training Details}
\subsection{Train Data}
To accomplish the tasks of visual navigation and video spatial understanding, we have collected a variety of datasets to enable more comprehensive training.

\subsubsection{Visual Navigation Task}
For the visual navigation task, we designed a supervised fine-tuning dataset (SFT data blend), which consists of the following four categories of data:  

\textbf{Navigation data from real videos}: By collecting trajectory videos from real-world scenes, we provide the model with continuous navigation examples in realistic environments.  

\textbf{Navigation data from simulation environments}: Using the R2R-CE and RxR-CE datasets, which provide sparse path point sequences converted from discrete VLN versions, we integrate these datasets into the Habitat simulator. This allows us to generate navigation video sequences based on the geodesic shortest path. These videos include consecutive frames and corresponding action labels, helping the model learn navigation capabilities in simulated environments.  

\textbf{Auxiliary navigation data}: We use augmented instructions generated by EnvDrop and introduce auxiliary tasks such as navigation trajectory summarization. Through these data, the model learns to describe trajectories across different time segments.  

\textbf{General VQA datasets}: To improve the model's generalization capabilities, we incorporate multiple general visual question answering (VQA) datasets. These datasets cover a wide range of scenes and real-world environments, providing the model with extensive training samples.  

\subsubsection{Video Spatial Understanding Task}
For the video spatial understanding task, we integrate multiple video question-answering datasets: 

\textbf{ScanQA dataset}: This dataset focuses on real-world 3D scene question-answering tasks, featuring free-form QA pairs grounded in 3D objects. Using this data, the model learns to reason and answer questions in 3D spaces.  

\textbf{Custom video QA dataset}: To further enhance the model's performance, we created custom datasets containing video data from various real-world scenarios. These datasets encompass diverse viewpoints, lighting conditions, and dynamic changes, providing the model with more challenging and diverse training samples.  

By integrating these datasets, we effectively enhance the model's capabilities in both navigation and video spatial understanding tasks, enabling it to perform well not only in simulation environments but also in real-world scenarios.

\subsection{Train Strategy}
To enhance the model's performance on video spatial understanding tasks, we fine-tuned the Qwen model using LoRA (Low-Rank Adaptation). The core idea of LoRA is to achieve parameter-efficient optimization by inserting trainable low-rank matrices without making large-scale adjustments to the original weights of the pre-trained model. This approach allows the model to adapt to downstream task requirements while maintaining its performance.

Specifically, LoRA decomposes the weight matrix $W \in \mathbb{R}^{d \times k}$ into two low-rank matrices $A \in \mathbb{R}^{d \times r}$ and $B \in \mathbb{R}^{r \times k}$. Through low-rank decomposition, the weight update formula can be expressed as $W' = W + A \cdot B$. During this process, the original weights $W$ remain frozen, and only the newly added low-rank matrices $A$ and $B$ are optimized. We inserted LoRA modules into key components of the model, including the query projection (\texttt{q\_proj}), value projection (\texttt{v\_proj}), and language model head (\texttt{lm\_head}), to enhance the model's ability to model dynamic relationships between video frames and perform cross-modal reasoning.

In the actual fine-tuning process, we set the LoRA hyperparameters as follows: rank $r=8$, LoRA scaling factor $\alpha=32$, and a Dropout probability of 0.1 to reduce the risk of overfitting. The LoRA modules were embedded into the model's multi-head self-attention mechanism and output layers, enabling more efficient adaptation to the requirements of video spatial understanding tasks. This method is particularly suitable for tasks involving multi-view video frames, such as reasoning about 3D spatial relationships from videos or answering scene-related questions. By incorporating LoRA, the model can better capture dynamic changes between video frames and improve its understanding of complex 3D scenes.

For video spatial understanding tasks, the use of LoRA allows us to efficiently optimize the model, making it more adaptable to downstream tasks while maintaining reasonable computational resource usage. This fine-tuning approach provides strong technical support for our video spatial understanding tasks.

\subsection{Hyperparameter Details}
\label{app:hyperparams}

To ensure the reproducibility of our results, we provide a comprehensive list of the key hyperparameters used for training the VEME model. We utilized the AdamW optimizer for stable and efficient training. The primary hyperparameters for the optimization and training process are summarized in Table~\ref{tab:hyperparams}. These settings were kept consistent across all main experiments unless otherwise specified in the ablation studies.

\begin{table}[ht]
    \centering
    \caption{Key hyperparameters for training the VEME framework.}
    \label{tab:hyperparams}
    \renewcommand{\arraystretch}{1.2} % Improves row spacing
    % --- 使用 resizebox 缩放表格 ---
    \resizebox{\linewidth}{!}{%
        \begin{tabular}{@{}ll@{}}
            \toprule
            \textbf{Hyperparameter} & \textbf{Value} \\
            \midrule
            \multicolumn{2}{l}{\textit{Optimizer \& Scheduler}} \\
            \quad Optimizer & AdamW \\
            \quad Learning Rate (Peak) & $1 \times 10^{-4}$ \\
            \quad Betas ($\beta_1, \beta_2$) & (0.9, 0.999) \\
            \quad Weight Decay & 0.01 \\
            \quad Learning Rate Schedule & Cosine decay \\
            \quad Warmup Steps & 500 \\
            \midrule
            \multicolumn{2}{l}{\textit{Training \& Batching}} \\
            \quad Total Training Epochs & 1 \\
            \quad Per-Device Batch Size & 8 \\
            \quad Gradient Accumulation Steps & 2 \\
            \quad Effective Batch Size & 64 ($8 \times 8 \text{ GPUs}$) \\
            \quad Mixed Precision & bfloat16 \\
            \midrule
            \multicolumn{2}{l}{\textit{Loss Configuration}} \\
            \quad Spatial Loss Weight ($\lambda_s$) & 0.1 \\
            \quad Episodic Loss Weight ($\lambda_e$) & 0.1 \\
            \quad Contrastive Temperature ($\tau$) & 0.07 \\
            \midrule
            \multicolumn{2}{l}{\textit{LoRA Fine-tuning}} \\
            \quad LoRA Rank ($r$) & 16 \\
            \quad LoRA Alpha ($\alpha$) & 32 \\
            \quad LoRA Dropout & 0.1 \\
            \quad LoRA Target Modules & \texttt{q\_proj}, \texttt{v\_proj}, \texttt{lm\_head} \\
            \bottomrule
        \end{tabular}%
    }
\end{table}

\paragraph{LoRA Configuration.}
As detailed in the `Train Strategy` subsection, we employed Low-Rank Adaptation (LoRA) for parameter-efficient fine-tuning. The rank ($r$) was set to 8, providing a good balance between expressiveness and parameter efficiency. The scaling factor ($\alpha$) was set to 32. To prevent overfitting on the adapter weights, a dropout rate of 0.1 was applied specifically to the LoRA modules. We targeted the query (\texttt{q\_proj}) and value (\texttt{v\_proj}) projections within the VLM's self-attention layers, as well as the final language model head (\texttt{lm\_head}), as these are critical for adapting the model's reasoning and generation capabilities to our specific tasks.

\section{Model Architecture Details}
\label{app:architecture_details}

Our proposed framework, VEME, is constructed around a central Vision-Language Model (VLM) and is augmented by a series of specialized encoders. Each encoder is designed to process a specific modality (vision, geometry, trajectory), transforming raw sensory input into rich feature representations. These features are then projected into a common embedding space, allowing the core VLM to perform holistic reasoning across all available information streams. Below, we detail the specifics of each key component.

\paragraph{Core VLM Backbone.}
The heart of our model is the \texttt{Qwen-2.5-VL-7B}~\cite{Bai2025Qwen25VLTR}, a powerful pretrained VLM that serves as our primary reasoning engine. We leverage its advanced capabilities in understanding and integrating vision and language. The input to this model is a carefully structured sequence of embeddings from various sources, as described in the main paper. We utilize the LoRA fine-tuning technique on its attention and feed-forward layers to adapt it to the downstream embodied tasks without catastrophic forgetting of its pretrained knowledge.

\paragraph{Visual Semantic Encoder.}
To extract high-level semantic information from each RGB frame $I_t$, we employ a pretrained \texttt{DINOv2-ViT-L/14} model~\cite{oquab2023dinov2}. \texttt{DINOv2} is renowned for its strong, self-supervised learned features that exhibit excellent performance on various downstream tasks without fine-tuning. For each frame, we extract the patch features from the final layer of the encoder. These features, which capture the objects and their appearance, are then passed through a two-layer Multi-Layer Perceptron (MLP) to project them from their native dimension to the VLM's hidden dimension $D$. This yields the semantic feature representation $F_{\text{vis}, t}$.

\paragraph{Image-based Geometric Encoder.}
Complementary to the semantic features, we extract immediate, view-dependent geometric cues directly from the 2D image $I_t$. For this, we use a specialized visual-geometry encoder, specifically a version of \texttt{VGGT-1B}~\cite{wang2025vggt} pretrained on multiple geometry estimation tasks. This model processes the RGB frame and outputs a dense feature map that implicitly encodes geometric information such as depth gradients and surface normals. These geometric features $F_{\text{geo}, t}$ provide the spatial context necessary to ground the semantic features $F_{\text{vis}, t}$ through our cross-attention mechanism. Like other features, these are also projected to dimension $D$ via an MLP.

\paragraph{3D Point Cloud Backbone.}
This is a critical component for building a persistent and global understanding of the environment's 3D structure. As mentioned in the main text, we use the \textbf{Sonata}~\cite{wu2025sonata} model as our dedicated 3D backbone.
\begin{itemize}
    \item \textbf{Architecture:} Sonata is a state-of-the-art sparse 3D Transformer architecture. It is specifically designed to efficiently process large-scale point clouds, which are common in real-world navigation scenarios. It utilizes a hierarchy of sparse voxel-based attention mechanisms, allowing it to capture both local geometric details and long-range spatial relationships without incurring the prohibitive computational cost of standard Transformers.
    \item \textbf{Input:} The model takes the global point cloud $P_t \in \mathbb{R}^{N_{pcd} \times 6}$ as input, where each point is represented by its XYZ coordinates and RGB color values.
    \item \textbf{Pre-training:} The Sonata backbone we use has been pretrained on a large-scale collection of 3D indoor scene datasets, including ScanNet~\cite{dai2017scannet} and Matterport3D~\cite{Matterport3D}, on a self-supervised reconstruction task. This pre-training endows it with a powerful, intrinsic understanding of 3D geometric primitives and typical spatial layouts of indoor environments.
    \item \textbf{Output:} The model outputs a set of per-point feature vectors $F_{\text{pcd}, t} \in \mathbb{R}^{N_{\text{pcd}} \times D}$ after being passed through a final projection layer. These features provide the foundational geometric representation for both our cognitive map and episodic memory modules.
\end{itemize}

\paragraph{Spatio-temporal Trajectory Encoder.}
To capture the agent's movement history, we encode the sequence of past actions $\mathcal{A}_{0:t-1}$. Each discrete action (e.g., `move\_forward', `turn\_left') is first mapped to a learnable embedding vector of dimension $D$. This sequence of action embeddings is then processed by a 4-layer Transformer encoder with 8 attention heads. The output of this encoder, $F_{traj,t}$, provides a contextualized representation of the agent's path, capturing temporal dependencies within the action sequence.

\section{Additional Experimental Results}

\subsection{Baseline details}
\label{app:benchmark_details} % 使用一个更具体的标签

\paragraph{Specialized Navigation Models}
For all specialized navigation baselines on VLN-CE, including \texttt{CMA}, \texttt{ETPNav}, and \texttt{NaVILA}, we adhered to the following protocol to ensure a fair and reproducible comparison:
\begin{itemize}
    \item \textbf{Source:} We utilized the official codebases and pre-trained model weights released by the respective authors. No architectural modifications were made.
    \item \textbf{Evaluation Protocol:} We followed the standard evaluation scripts and environment setups provided with each baseline's repository. Results were generated on the \texttt{val-unseen} split as reported in the original papers.
\end{itemize}

\paragraph{General Vision-Language Models}
For all general-purpose VLMs evaluated on VSI-Bench, such as \texttt{GPT-4o}, \texttt{Gemini-1.5 Pro}, the \texttt{LLaVA} series, and the \texttt{Qwen} series, we used the following zero-shot evaluation setup:
\begin{itemize}
    \item \textbf{Model Version:} We used the latest available official APIs for proprietary models (e.g., \texttt{gpt-4o-2024-05-13}) and the official Hugging Face implementations for open-source models.
    \item \textbf{Prompting Strategy:} A consistent, minimal prompt template was employed across all models to query their spatial reasoning capabilities without providing few-shot examples. The templates were structured as follows:

    \begin{quote}
    \small
    \textbf{For Multiple-Choice Questions:}\\
    \texttt{The following is a video of an indoor scene. Based on the video, answer the following question by choosing the best option.
    <video\_placeholder>
    Question: [Question from VSI-Bench]
    Options:
    (A) [Option A]
    (B) [Option B]
    (C) [Option C]
    (D) [Option D]
    Answer (Provide the letter only):}
    \end{quote}
    
    \begin{quote}
    \small
    \textbf{For Numerical Questions:}\\
    \texttt{The following is a video of an indoor scene. Based on the video, answer the following question. Provide only the numerical value in your answer.
    <video\_placeholder>
    Question: [Question from VSI-Bench]
    Answer:}
    \end{quote}
\end{itemize}

\subsection{Computational Cost and Efficiency}

\subsection{Evaluation Protocol Details}

\begin{itemize}
    \item \textbf{Metrics Calculation:} All metrics were calculated using the official evaluation scripts provided by the respective benchmarks. For VLN-CE, we report Success Rate (SR) and Success rate weighted by Path Length (SPL). For VSI-Bench, we report Accuracy (ACC) for multiple-choice questions and Mean Relative Accuracy (MRA) for numerical questions.
    \item \textbf{Inference Setup:} All our models were evaluated on a single NVIDIA A100 GPU. We used greedy decoding (i.e., beam size of 1) for generating all responses to ensure efficiency and deterministic outputs. All reported scores are the average of three evaluation runs with different random seeds to ensure statistical stability.
\end{itemize}

\subsection{Ablation Experiment}
To assess the significance of performance changes in our ablation studies, we used the Wilcoxon signed-rank test, which is ideal for small sample sizes (n=5) and does not rely on specific data distribution assumptions. The null hypothesis posits that the median difference between paired observations is zero. If we reject the null hypothesis ($p < 0.01$), it indicates a statistically significant performance drop after removing certain modules.

We calculated the performance metrics for different configurations, including the full model and various ablations. Our analysis revealed significant performance degradation across all removed components. Notably, the removal of episodic memory resulted in the largest decrease ($Delta=-15.3\%$, $p = 0.0003$), followed by trajectory input ($\Delta = -12.9\%$, $p = 0.0005$), spatial memory ($\Delta = -9.8\%$, $p = 0.0007$), and geometric grounding ($\Delta = -3.9\%$, $p = 0.0011$). All $p$-values were below the $0.01$ threshold, underscoring the critical roles these components play in overall system performance.

\begin{figure*}[ht!]
    \centering
    % --- 请将此处的 \fbox 替换为你的 t-SNE 可视化图片 ---
    \includegraphics[width=\textwidth]{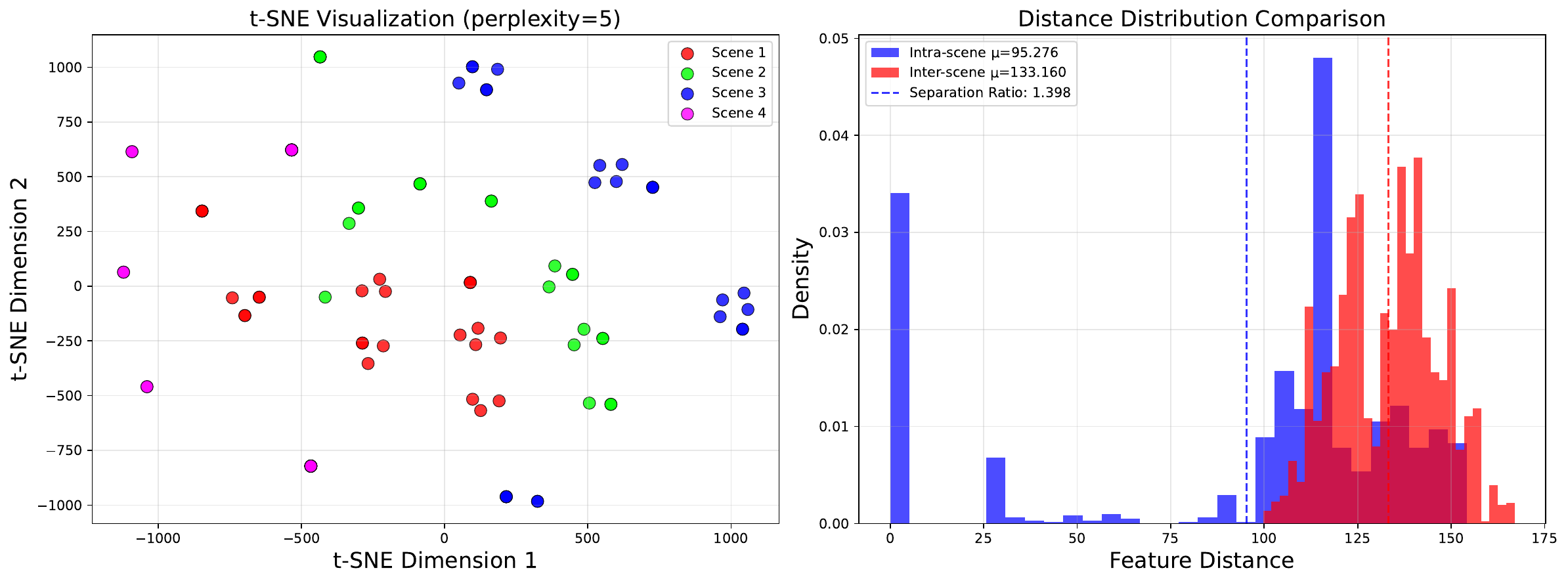}
    \caption{\textbf{t-SNE and distance distribution visualization of episodic features ($F_{\text{episodic}}$).} \textbf{Left:} Each point corresponds to a full navigation episode. Points are colored by their environment ID. The clear clustering of points of the same color demonstrates that our model learns discriminative representations for different spatio-temporal experiences. \textbf{Right:} \textcolor{blue}{Blue} distribution represents intra-scene episodic feature distances, while \textcolor{red}{Red} distribution represents inter-scene ones. The separation of episodic features demonstrates effective experience learning in our model.}
    \label{fig:tsne_episodes}
\end{figure*}

In a similar analysis for spatio-temporal understanding, we found that removing the vggt component led to significant drops across all metrics, while fine-tuning removal produced slightly stronger effects. The dramatic reduction in mean recognition accuracy (MRA) further emphasizes the importance of these components for effective temporal reasoning, with all $p$-values also falling below $0.01$.

\subsection{Additional Qualitative Results}
To provide a more comprehensive understanding of our model's capabilities and limitations, we present additional qualitative examples below.

\paragraph{Success Case}
The main text has already highlighted our model's ability to follow long and complex instructions, successfully navigating through multiple rooms, while a strong baseline fails by becoming stuck in a local loop.

\begin{figure}[h!]
    \centering
    \includegraphics[width=\linewidth]{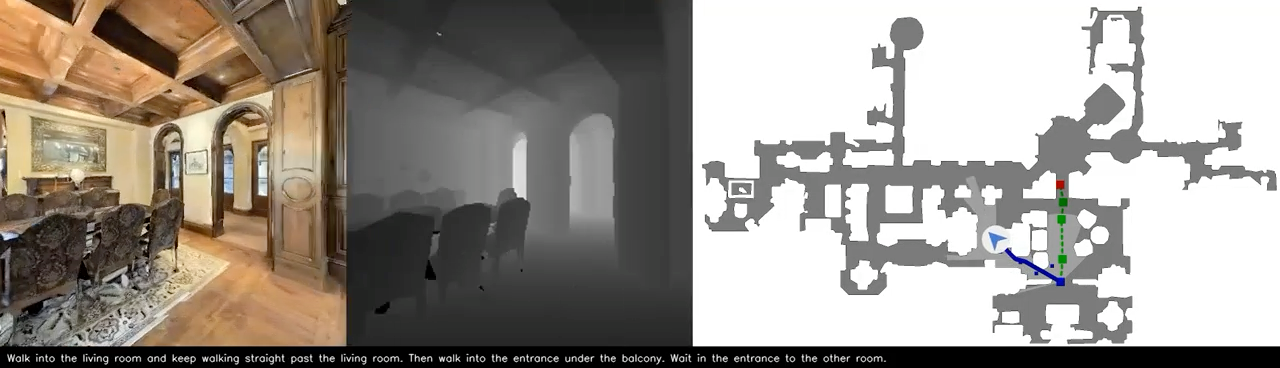}
    \caption{Failure case illustration with the instruction: "Walk into the living room and keep walking straight past the living room. Then walk into the entrance under the balcony and wait at the entrance to the other room."}
    \label{fig:failure_case}
\end{figure}

\paragraph{Failure Case Visualization and Analysis}
Our model has its limitations, as illustrated in Figure~\ref{fig:failure_case}. This figure presents a failure case with the instruction: "Walk into the living room and keep walking straight past the living room. Then walk into the entrance under the balcony and wait at the entrance to the other room." Despite the clear visual cues provided in the left image, the model struggles to accurately interpret the spatial layout, leading to its failure to select the optimal path.

\section{Feature Visualization}

To provide a more intuitive understanding of our model's internal mechanisms and validate its claimed capabilities, we present a series of qualitative visualizations. These analyses aim to demystify how our model learns discriminative spatio-temporal representations, grounds semantics in geometry, and reasons about complex, instruction-driven tasks.

% First, 
We sought to verify that our Episodic Memory module creates unique and separable representations for different spatio-temporal experiences. To visualize this, we sampled a collection of distinct navigation episodes from the validation set and extracted the final episodic feature vector, $F_{\text{episodic}}$, for each. Using the t-SNE dimensionality reduction algorithm, we projected these high-dimensional vectors into a 2D space. The result, shown in Figure \ref{fig:tsne_episodes}, is a scatter plot where each point represents a complete episode, colored by its environment ID, and its corresponding distance distribution. The clear clustering of points with the same color provides strong evidence that our model, guided by the $\mathcal{L}_{\text{episodic}}$ loss, successfully learns to form a discriminative memory space. This ability to distinguish between different journeys is fundamental for long-term reasoning and avoiding ambiguity.

\section{Limitations and Future Work}
\label{sec:limitations_future_work}

While our proposed VEME framework demonstrates significant advancements in equipping embodied agents with spatio-temporal intelligence, we acknowledge several limitations that pave the way for compelling future research.

First, our model's multi-encoder architecture, while powerful, introduces a notable computational overhead in terms of both memory (VRAM) and floating-point operations (FLOPs) during inference. This currently may hinder its deployment on resource-constrained robotic platforms with limited onboard computing power. Second, the performance of our geometric and episodic memory modules is, to some extent, contingent on the availability of relatively clean 3D point clouds and accurate trajectory data during the training phase. The model's robustness against noisy, incomplete, or drifting map data generated by real-world SLAM systems is an important area for further investigation. Finally, our current framework operates under a static environment assumption. It is not designed to handle dynamic scenes with moving objects, interacting agents, or significant changes in layout during an episode, which limits its applicability in more complex, human-centric settings.

Addressing these limitations points toward several exciting future directions. A primary focus will be on \textbf{improving model efficiency}. We plan to explore model compression techniques, such as knowledge distillation from our larger model to a more compact student network, and post-training quantization to create a "VEME-Lite" version without substantially compromising performance. This would be a critical step towards real-world robotic deployment. 

Another key avenue is to \textbf{enhance the model's robustness and generalization}. To bridge the sim-to-real gap highlighted by our data dependency, we aim to train VEME on a much wider variety of 3D environments, incorporating simulated sensor noise and diverse visual styles. Furthermore, we will investigate advanced learning paradigms, such as meta-learning or online domain adaptation, to enable the agent to rapidly fine-tune its world model when introduced to a completely new and unseen environment.

Perhaps the most significant long-term vision is to \textbf{transcend the offline training paradigm towards interactive and lifelong learning}. This involves developing mechanisms for the agent to continuously update its semantic and episodic memories based on its own experiences and interactions within the world. By integrating reinforcement learning principles, the agent could learn from trial-and-error, associate failed plans with specific environmental states, and implicitly refine its understanding of physical affordances. Such a system would move us closer to creating truly adaptive and autonomous embodied intelligences that learn and grow over their entire operational lifetime.

% \begin{figure}[h!]
%     \centering
%     % 请替换为你的失败案例分析图
%     % \includegraphics[width=0.8\linewidth]{path/to/your/failure_case_figure.pdf} 
%     \fbox{\parbox[c][15em][c]{0.8\linewidth}{\centering Placeholder for Failure Case Analysis Figure}}
%     \caption{\textbf{Failure Case Analysis.} The instruction is "Go to the chair on the left." In this highly symmetrical room with two identical chairs, our model hesitates between the two and ultimately chooses the incorrect one. This highlights a limitation in resolving fine-grained spatial ambiguity without additional context.}
%     \label{fig:failure_case}
% \end{figure}

\clearpage

\end{document}